\documentclass{llncs}
\usepackage{llncsdoc}
\usepackage{graphics} 
\usepackage{epsfig} 
\usepackage{mathptmx} 
\usepackage{enumerate} 
\usepackage{amsmath} 
\usepackage{amssymb}  
\usepackage{subfigure}
\usepackage{caption}
\usepackage{url}
\begin{document}

\title{Optimizing Cost-Sensitive SVM for Imbalanced Data :Connecting Cluster to Classification}


\author{Qiuyan Yan , Shixiong Xia , Fanrong Meng }

\authorrunning{} 
\institute{School of Computer Science and Technology,\\
China University of Mining Technology,Xuzhou 221116, China}

\date{Received: date / Accepted: date}

\maketitle
\begin{abstract}
Class imbalance is one of the challenging problems for
machine learning in many real-world applications, such as coal and gas burst
accident monitoring: the burst premonition data is extreme smaller than the
normal data, however, which is the highlight we truly focus on.
Cost-sensitive adjustment approach is a typical algorithm-level method
resisting the data set imbalance. For SVMs classifier, which
is modified to incorporate varying
penalty parameter(C) for each of
considered groups of examples. However, the C value is determined empirically, or is
calculated according to the evaluation metric, which need to be computed
iteratively and time consuming. This paper presents a novel cost-sensitive SVM method whose penalty parameter C optimized on the basis of cluster probability density function(PDF) and the cluster PDF is estimated only according to similarity matrix and some predefined
hyper-parameters. Experimental results on various standard benchmark
data sets and real-world data with different ratios of imbalance show that
the proposed method is effective in comparison with commonly used
cost-sensitive techniques.
\end{abstract}

\section{Introduction}

Support vector machines (SVMs) is a popular machine learning technique,
owning to its theoretical and practical advantages, which has been
successfully employed in many real-world application domains. However, when
shifting to imbalanced dataset, SVM will produce an undesirable model that
is biased toward the majority class and has low performance on the minority
class. Imbalanced learning not only presents significant new challenges to the data
research community~\cite{1,2,3} but also raises many critical questions in real-world data intensive applications.In fact, in most imbalanced learning problems, the
misclassification error of the minority class is far costlier than that of
the majority class. Following we will give
an example to clarify this problem.

\begin{figure}[h]
\centering
\includegraphics[width=.76\textwidth]{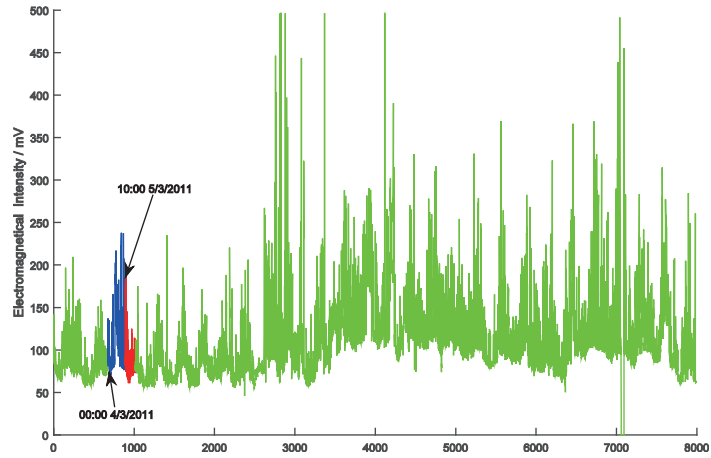}
\caption{Mine and Gas Burst Electromagnetic Monitoring Data (electromagnetic
intensity).The red line means taking place mine and gas burst accident on the date of 05/03/2011.
The blue line means the burst premonition data.Compared to normal data,premonition data is extreme small, but which is the object we really focused on.  }
\label{fig1}
\end{figure}

Figure~1 is the mine and gas burst electromagnetic monitoring data for one
month. The red line means taking place mine and gas
burst accident on the date of 05/03/2011. One day before the accident, the
blue line, means the burst premonition data, which is the object we focused on.
If we can classify correctly the premonition data from the other normal
data, we can pre-alarm the burst accident and avoid the accident taking
place. From the Fig.~\ref{fig1}, one month electromagnetic monitoring data is 7989,
and there are only 210 premonition data. If we set the premonition data is
positive class, and the other normal data is negative class, then the
Imbalance Ratio, equals to the ratio of negative class data number and
positive class data number, that is 37.04. In fact, the mine and gas burst
accident rarely take place, so the imbalance ratio even larger. For this
burst electromagnetic monitoring data, we expect positive class and negative
class can be classified correctly under this extreme imbalance ratio.

SVMs classifier resist the data imbalance from three categories: Data level,
Algorithm level and Hybrid methods. \textbf{Data-level methods} mainly
includes oversampling and undersampling, which may have the danger of hardly
maintain the original minor data distribution and also may lose the
important majority data information seperately. \textbf{Algorithm-level methods}
concentrate on modifying existing learners to alleviate their bias towards
majority groups. Cost-sensitive methods is a typical representation of this
category. The main problem is it is difficult to set the actual values in
the cost matrix and often they are not given by expert beforehand in many
real-life problems. The penalty parameter(C)
is determined empirically, or
is calculated according to the
evaluation metric, which need
to be computed iteratively and
time consuming. Decision threshold adjustment is another kind of
algorithm-level solution for dealing with class imbalance. Unlike the other
correction techniques, decision threshold adjustment strategy runs after
modeling a classifier. However, the existing decision threshold adjusting
approaches generally give the moving distance of classification boundary
empirically, but which cannot answer the question that how far the
classification hyperplane should be moved towards the majority class.
\textbf{Hybrid methods} usually combine the algorithm-level methods and
data-level methods to extract their strong points and reduce their weakness,
which can be applied to arbitrary classifier and not only restrict to SVMs.

From the other side, recently, one interesting directions indicates that
imbalance ratio is not the sole source of learning difficulties. Even if the
disproportion is high, but both classes are well represented and come from
non-overlapping distribution we may obtain good classification rates using
canonical classifiers. Nevertheless, a good imbalance learning algorithm
should fully understand and exploit the minority class structure.

The cluster probability density function(PDF) can naturedly express the class structure and distribution.In this paper, we proposed an cluster \textbf{P}robability optimized \textbf{CS-SVM} method ,named as \textbf{PCS-SVM} ,whose C value determined by the cluster probabilistic density function.We adjust the hyperplane through optimization the value of C
for the positive class data, that is adjustment the upper bound of $\alpha
$. First, we introduce the L2 norm to optimization function, the dual
Lagrangian form of the modified objective function includes the parameter of
$C^{+}$ and $C^{-}$. Compared to the traditional SMO algorithm results, the
new $\alpha $ expression function is only different from the similarity
expression of support vector $x$ and $y$. Then, we construct the connection
between similarity and the cluster probability of two points. This is an
important measure differentiate the probability density of majority class
and minority class. Last, selection two opposite support vector, one is
positive but false classified to negative, the other is true negative. By
correcting a false negative support vector to a true positive support vector
according to cluster possibility, achieve the decision boundary removing
toward the majority (negative) class.

\section{Related works}

In previous work, the class imbalance correction strategies for SVM mainly
includes resampling~\cite{4} ,cost-sensitive~\cite{5,6}, decision threshold adjustment~\cite{7}, and
hybrid methods~\cite{8}. Resampling can be accomplished either by
oversampling the minority class or undersampling the majority class.
However, both sampling techniques have their advantages and disadvantage.
Oversampling makes the classifier overfitting and increases the time of
modeling, while undersampling often causes information loss.In order to bypass this problem,~\cite{4}
proposed a Model-Based Oversampling method for imbalanced time series data
considering the sequence structure when oversampling. Considering data structure and distribution is the trend of this area.

The Different Error Costs (\textit{DEC}) method is a cost-sensitive learning solution
proposed in~\cite{9} to overcome the same cost (i.e. C) for both
positive and negative misclassification in the penalty term. As given in
Equation~(1)
\begin{align}
\label{eq1}
&\min (\frac{1}{2}\left\| w \right\|^{2}+C^{+}\sum\limits_{y_{i} =+1} {\xi
_{i}} +C^{-}\sum\limits_{y_{i} =-1} {\xi_{i}} )\\
&s.t. \begin{cases}
 y_{i} -(w^{T}x_{i} +b)\geqslant 1-\xi_{i} \\
 \xi_{i} \geqslant 0 \\
 \end{cases}\nonumber
\end{align}
In this method, the SVM soft margin objective function is modified to assign
two misclassification costs, $C^{+}$ and $C^{-}$ is the misclassification cost
for positive and negative class examples separately. As a rule of thumb, \cite{10} have reported that reasonably good classification results could
be retained from the DEC method by setting the ${C^{-}}/{C^{+}}$ equal to
the minority-to-majority class ratio. One-class Learning~\cite{11} trained an SVM model only with the minority class examples. ~\cite{12} assigned $C^{-}=0$ and $C^{+}=1/{N^{+}}$, these methods have
been observed to be more effective than general data rebalancing methods.
\textbf{\textit{z}}\textit{SVM} is another algorithm modification proposed for SVMs in~\cite{13} to learn from imbalance datasets, which is an typical decision
threshold adjustment method. In this method, first an SVM model is developed
by using the original imbalanced training dataset. Then, the decision
boundary of the resulted model is modified to remove its bias toward the
majority class. In zSVM method, the magnitude of the $\alpha_{i}^{+}$ is
increased by multiplying all of them by a particular value $z$. Then, the
modified SVM decision function can be represented as follows:
\begin{equation}
\label{eq2}
f(x) ={\rm sign}(z\ast \sum\limits_{i=1}^{N_{1}} {\alpha_{i}
^{+}y_{i} K(x_{i}, x)+\sum\limits_{i=1}^{N_{2}} {\alpha_{i}^{-}y_{i}
K(x_{i}, x)+} b)}
\end{equation}
The $z$ value is optimized by gradually increasing the value of $z$ from 0 to
some positive value, $M$, and G-mean is adopted as the evaluation measure to
determine the optimal $z$. The speed and efficiency of zSVM determined by the
step length from 0 to $M$.

\section{Proposed method}

According to DEC method, the optimized objective function has two loss
terms:
\begin{align}
\label{eq3}
&\min (\frac{1}{2}\left\| w \right\|^{2}+C^{+}\sum\limits_{y_{i} =+1} {\xi
_{i}} +C^{-}\sum\limits_{y_{i} =-1} {\xi_{i}} )\\
&s.t. \begin{cases}
 y_{i} -(w^{T}x_{i} +b)\geqslant 1-\xi_{i} \\
 \xi_{i} \geqslant 0 \\
 \end{cases}\nonumber
\end{align}
If introduce L2 norm regularization item of slack factor, Equation~(3) can
be converted into the Equation~(4):
\begin{align}
\label{eq4}
 &\frac{1}{2}\left\| w \right\|^{2}+C^{+}\sum\limits_{y_{i} =+1} {\xi
_{i}^{2}} +C^{-}\sum\limits_{y_{i} =-1} {\xi_{i}^{2}} \\
& s.t. \begin{cases}
 y_{i} -(w^{T}x_{i} +b)\geqslant 1-\xi_{i} \\
 \xi_{i} \geqslant 0 \\
 \end{cases} \nonumber
\end{align}
The dual Lagrangian form of the modified objective function is:
\begin{align}
L_{p} &=\sum\limits_{i=1}^p {\alpha_{i}} -\frac{\left\| w
\right\|^{2}}{2}+C^{+}\sum\limits_{\{i|y_{i} =+1\}}^p {\xi_{i}^{2}}
+C^{-}\sum\limits_{\{i|y_{i} =-1\}}^p {\xi_{i}^{2}} \nonumber\\
&\quad~-\sum\limits_{i=1}^p
{\alpha_{i}} [y_{i} (w\cdot x_{i} +b)-1+\xi_{i}
]-\sum\limits_{i=1}^p {\mu_{i} \xi_{i}}
\end{align}
Where $\alpha_{i} \geqslant 0$, and $\mu_{i} \geqslant 0$.
Partial deviation formula (5):
\begin{align}
\label{eq5}
&\frac{\partial L}{\partial w}=\left\| w \right\|-\sum\limits_{i=1}^p {\alpha
_{i} y_{i} x_{i}} =0\Rightarrow \left\| w
\right\|=\sum\limits_{i=1}^p {\alpha_{i} y_{i} x_{i}}\\
\label{eq6}
&\frac{\partial L}{\partial b}=\sum\limits_{i=1}^p {\alpha_{i} y_{i}} =0
\\
\label{eq7}
&\frac{\partial L}{\partial \xi_{i}} =C_{i}^{+} \sum\limits_{y_{i} =+1} {\xi
_{i}} +C_{i}^{-} \sum\limits_{y_{i} =-1} {\xi_{i}} -\sum\limits_{i=1}^p
{\alpha_{i}} -\sum\limits_{i=1}^p {\mu_{i} =0}
\end{align}
Substituting (6)--(8) into (5) yields the following:
\begin{align}
\label{eq8}
 L_{p} &=\sum\limits_{i=1}^p {\alpha_{i}} -\frac{\left\| w
\right\|^{2}}{2}+C^{+}\sum\limits_{\{i|y_{i} =+1\}}^p {\xi_{i}^{2}}
+C^{-}\sum\limits_{\{i|y_{i} =-1\}}^p {\xi_{i}^{2}} \nonumber\\
&\quad~-\sum\limits_{i=1}^p
{\alpha_{i}} [y_{i} (w\cdot x_{i} +b)-1+\xi_{i}
]-\sum\limits_{i=1}^p {\mu_{i} \xi_{i}} \nonumber\\
& =\sum\limits_{i=1}^p {\alpha_{i}}
-\frac{1}{2}\sum\limits_{i=1}^p {\alpha_{i} \alpha_{j} y_{i} y_{j} x_{i}
^{T}x_{j}} +C^{+}\sum\limits_{\{i|y_{i} =+1\}}^p {\xi_{i}^{2}}
+C^{-}\sum\limits_{\{i|y_{i} =-1\}}^p {\xi_{i}^{2}} -\sum\limits_{i=1}^p
{(\alpha_{i} +\mu_{i})\xi_{i}}
\end{align}
According to KKT condition, $\sum\limits_{i=1}^p {\mu_{i} \xi_{i}} =0$,
for $\xi_{i} \ne 0$, so $\mu_{i} =0$. Based on Equation~(9), we can get
\begin{equation}
\label{eq9}
\sum\limits_{i=1}^p {\alpha_{i}} =2C^{+}\sum\limits_{y_{i} =+1}^p {\xi
_{i}} +2C^{-}\sum\limits_{y_{i} =-1}^p {\xi_{i}}
\end{equation}
Substituting (10) into (9) yields the following
\begin{align}
\label{eq10}
 L_{p} &=\sum\limits_{i=1}^p {\alpha_{i}} -\frac{1}{2}\sum\limits_{i=1}^p
{\alpha_{i} \alpha_{j} y_{i} y_{j} x_{i}^{T}x_{j}}
-C^{+}\sum\limits_{\{i|y_{i} =+1\}}^p {\xi_{i}^{2}}
-C^{-}\sum\limits_{\{i|y_{i} =+1\}}^p {\xi_{i}^{2}} \nonumber\\
 &=\sum\limits_{i=1}^p {\alpha_{i}}
-\frac{1}{2}\sum\limits_{i=1}^p {\alpha_{i} \alpha_{j} y_{i} y_{j} x_{i}
^{T}x_{j}} -\frac{\sum\limits_{\{i|y_{i} =+1\}}^p {\alpha_{i}^{2}}
}{4C^{+}}-\frac{\sum\limits_{\{i|y_{i} =-1\}}^p {\alpha_{i}^{2}}} {4C^{-}}
\end{align}
According to SMO algorithm, the original object function including L1 norm
regularization item of slack factor for Equation~(3) can be described as:
\begin{align}
\label{eq11}
 \alpha _{2}^{new} &=\frac{y_{2} [y_{2} -y_{1} +y_{1} \gamma
(K_{11} -K_{12})+v_{1} -v_{2}]}{K_{11} +K_{22} -2K_{12}} \nonumber\\
 &=\alpha _{2}^{old} +\frac{y_{2} (E_{1} -E_{2})}{K},
\end{align}
Where $E_{i} =f(x_{i})-y_{i} , K=K_{11} +K_{22} -2K_{12} $.
$
K_{ij} =\sum\limits_{i=1}^p {x_{i}^{T}x_{j}} , f(x_{i})=\sum\limits_{i=1}^p {\alpha_{i} y_{i} K_{ij}} +b , v_{i} = f(x_{i})-\sum\limits_{j=1}^2 {\alpha
_{j} y_{j} K_{ij} -b}.
$
When $L_{p} =\sum\limits_{i=1}^p {\alpha_{i}}
-\frac{1}{2}\sum\limits_{i=1}^p {\alpha_{i} \alpha_{j} y_{i} y_{j} x_{i}
^{T}x_{j}} -\frac{\sum\limits_{\{i|y_{i} =+1\}}^p {\alpha_{i}^{2}}
}{4C^{+}}-\frac{\sum\limits_{\{i|y_{i} =-1\}}^p {\alpha_{i}^{2}}
}{4C^{-}}$, we select two opposite symbol support vector to modify $L_{p}
 $ expression: One is false negative support vector, the other is
true positive support vector. Under this condition, given $\alpha_{1}
 $ is negative and $\alpha_{2} $ is positive, the
optimization function turns into:
\begin{align}
L(\alpha_{2})&=\gamma -s\alpha_{2} +\alpha_{2} -\frac{1}{2}(\gamma
-s\alpha_{2})^{2}K_{11} -\frac{1}{2}\alpha_{2}^{2} K_{22} -s(\gamma
-s\alpha_{2})\alpha_{2} K_{12} \nonumber\\
&\quad~-y_{1} (\gamma -s\alpha_{2})v_{1} -y_{2}
\alpha_{2} v_{2} +constant-\frac{\alpha_{1}^{2}} {4C^{-}}-\frac{\alpha
_{2}^{2}} {4C^{+}}
\end{align}

Then in order to simplicity, we give $C^{-}=1$, and due to $\alpha_{1}
=\gamma -s\alpha_{2} $, $s=y_{1} \ast y_{2} =-1$, derivate of Equation~(13), the results is
\begin{align}
 &-s+1+s\gamma K_{11} -K_{11} \alpha_{2} -\alpha_{2} K_{22} -s\gamma K_{12}
+2\alpha_{2} K_{12} +sy_{1} v_{1} -y_{2} v_{2} -\frac{\alpha_{2}
}{2C^{+}}-\frac{\alpha_{2}} {2}=0 \nonumber\\
 &\Rightarrow \alpha_{2}^{new} =\frac{y_{2} (E_{1} -E_{2})+\alpha
_{2}^{old} (K_{11} +K_{22} -2K_{12})}{K_{11} +K_{22} -2(K_{12}
-\dfrac{1}{4C^{+}}-\dfrac{1}{4})}
 \end{align}

Compared to Equation~(11), the expression of $\alpha_{2}^{new} $ has the
change in denominator:
\begin{align*}
\text{From $K_{11} +K_{22} -2K_{12} $ to $K_{11} +K_{22} -2(K_{12}
-\frac{1}{4C^{+}}-\frac{1}{4})$}
\end{align*}

From the Equation~(14), we can conclude that considering the cost of false
positive classification can by means of adjusting the value of similarity
matrix, that is, update the similarity of $K_{12} $ to $K_{12}^{new}
=K_{12}^{old} -\frac{1}{4C^{+}}-\frac{1}{4}$. Through this adjustment, the
new value of $\alpha_{2} $ will decrease, under the condition of the
other $\alpha $ unchanged, $\left\| w \right\|=\sum\limits_{i=1}^p {\alpha
_{i} y_{i} x_{i}} $ will also decrease, that is, the max margin of
hyperplane will increase. As the positive support vector we selected are
unchanged, increasing of max margin means the hyperplane move towards the
majority data. Then, we will discuss how to determine the value
of $K_{12}^{new} $.

\begin{figure}[h]
\centering
\subfigure[Before adjustment the hyperplane position. False negative support vector A and true positive support vector B belong to different class according to the similarity of A and B]{\includegraphics[width=.486\textwidth]{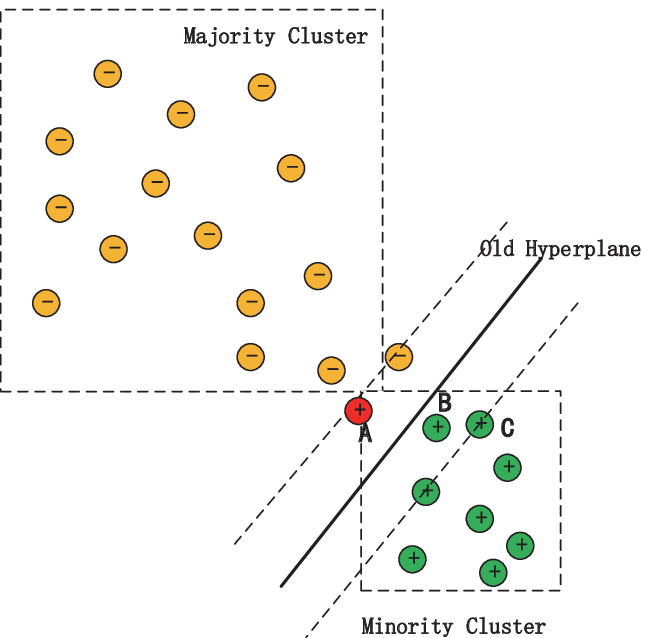}}\hfill
\subfigure[After adjustment the hyperplane position. Adjusting similarity of A and B to the similarity of B and C, the new slack factor C+ is optimized and the new hyperplane is moved toward the majority class]{\includegraphics[width=.486\textwidth]{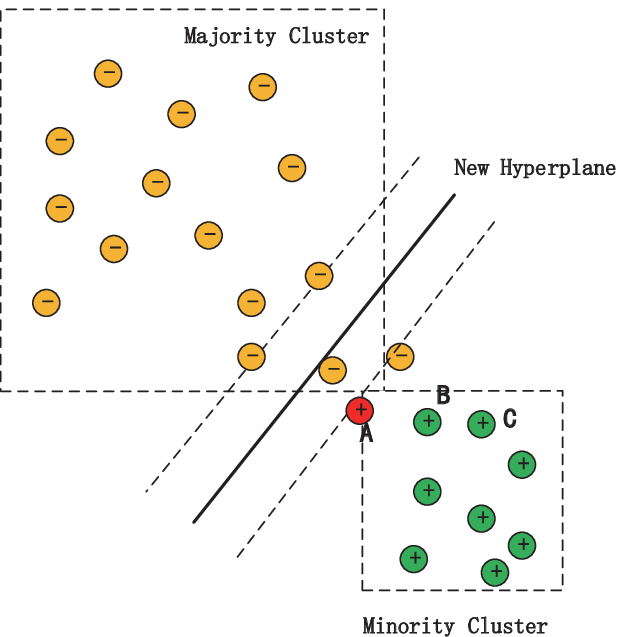}}
\caption{The strategy on how to move the hyperplane toward majority class}
\label{fig2}
\end{figure}

In Fig.~2, green points are the minority (positive) class, and the yellow
points are the majority (negative) class. At the first, in order to maintain
the correctness of majority class, the hyperplane is bias against the
minority, whereas the false negative rate is high. Based on the first
classifier model, we select two nearest support vectors, one is false
negative (point A, in red color), and the other is true positive (point B,
in green color). For the old hyperplane (shown as in Fig.~2(a)), point A is
more near to the negative point according to the similarity function than
the other positive ones. If we can adjust the similarity of point A and its
nearest true positive neighbor (such as point B), let the similarity larger
than the similarity of point B and its nearest true positive neighbor (such
as point C), then the new hyperplane (shown as in Fig.~2(b)) will classify the
point A correctly.

The next problem is how to determine the new $K_{12}^{new} $. We only know
that using the old learning model, A and B are classified as different
class, and B and C are classified as the same class. If we suppose the
probabilistic density function of minority cluster is $P_{1} (K_{ij} )$, and that of different cluster is $P_{0} (K_{ij} )$, then
the probability of B and C belong to the minority cluster is $p_{BC} =P_{1}
(K_{BC} )$, and the probability of A and B belong to the
different cluster is $p_{AB} =P_{0} (K_{AB} )$, the probability of
A and B belong to the minority cluster is $p_{AB}^{new} =P_{1} (K_{AB}
 )$. Then we can get the Equation~(15)
\begin{equation}
\label{eq12}
\frac{K_{AB}} {K_{AB}^{new}} \propto
\frac{p_{AB}} {p_{AB}^{new}} ,~~
\frac{K_{AB}} {K_{BC}} \propto \frac{p_{AB}
} {p_{BC}}
\end{equation}
Under the same probabilistic density function $P_{1} (K_{ij} )$,
if we hope $p_{AB}^{new} >p_{BC} $, then $K_{AB}^{new} >K_{BC} $, we can get:
\begin{align}
\label{eq13}
&\frac{K_{AB}} {K_{AB}^{new}} \leq
\frac{K_{AB}} {K_{BC}} \propto \frac{p_{AB}
} {p_{BC}} \Rightarrow K_{AB_{new}} \geq
k\times K_{AB} \times \frac{p_{BC}} {p_{AB}}\\
& K_{AB}^{new} =K_{AB} -\frac{1}{4C^{+}}-\frac{1}{4}>k\times K_{AB} \times
\frac{p_{BC} }{p_{AB} } \nonumber\\
& \Rightarrow K_{AB} (1-k\times \frac{p_{BC} }{p_{AB} })-\frac{1}{4}>\frac{1}{4C^{+}} \nonumber\\
& \Rightarrow K_{AB} \cdot \frac{p_{AB} -k\times p_{BC} }{p_{AB}
}-\frac{1}{4}>\frac{1}{4C^{+}} \nonumber\\
 &\Rightarrow C^{+}>\frac{p_{AB} }{4K_{AB} (p_{AB} -k\times p_{BC})-p_{AB} }\nonumber
\end{align}
Without loss of generality, we set $k=1$. In Equation~(16), the value of
$C^{+}$ depends on three parameters, $K_{AB},p_{AB} $ and $p_{BC} $. $K_{AB}
$ is known, so the problem is how to obtain the $p_{AB} $ and $p_{BC}
$. According to the reference ~\cite{14}, the similarity matrix
${W}$ is modeled as beta distributions, which are defined on the
interval of (0, 1), parameterized by two positive shape parameters $\alpha
$ and $\beta $. In this case, let $W_{ij} $ is a sample from a beta
distribution that is parameterized by $\Theta_{k} =(\alpha_{k}, \beta_{k}
)$, such that it is skewed towards one. If $W_{ij} $ is in off-diagonal
blocks, then let $\Theta_{0} =(\alpha_{0}, \beta_{0})$ be the parameters
for the beta distribution, then it is smaller than any other beta
distributions. The probability density function of $W_{ij} $ can
be expressed as:
\begin{equation}
\label{eq15}
p(W_{ij} |\{\Theta_{k} \}_{k=1}^{K}, \Theta_{0}, Z)=Beta(W_{ij} |\alpha
_{0}, \beta_{0})^{1-\sum\nolimits_k^K {z_{ik} z_{jk} }
}\prod\limits_{k=1}^K {Beta(W_{ij} |\alpha_{k}, \beta_{k})^{z_{ik} z_{jk}}}
\end{equation}
$z_{n} $ is a K-element cluster indicator $z_{n} =\{z_{nk} \}_{k=1}^{K} $ such
that $z_{nk} =1$ if $x_{n} $ belongs to the $k$-th cluster, and otherwise $z_{nk}
=0$. $z_{n} $ follows a categorical distribution, and $\pi $ is a sample from a
symmetric Dirichlet distribution

The prior distributions to the beta distribution parameters $\Theta_{k}
$ and $\Theta_{0} $ are given:
\begin{align}
\label{eq16}
&p\{\Theta_{k} |\varsigma)\infty Beta(\frac{\alpha_{k} }{\alpha_{k}
+\beta_{k} }|\alpha_{\varsigma }, \beta_{\varsigma })Lognormal(\alpha
_{k} +\beta_{k} |\mu_{\varsigma }, \sigma_{\varsigma }^{2})
\\
\label{eq17}
&p\{\Theta_{0} |\eta)\infty Beta(\frac{\alpha_{k} }{\alpha_{k} +\beta
_{k} }|\alpha_{\eta }, \beta_{\eta })Lognormal(\alpha_{\eta } +\beta
_{\eta } |\mu_{\eta }, \sigma_{\eta }^{2})
\end{align}
We want to calculate the posterior distribution for the latent variables
given the observed similarity matrix and the hyper-parameters, i.e.
\begin{align}
p(\pi, Z, \{\Theta_{k} \}_{k=1}^{K}, \Theta_{0} |W, \varsigma, \eta, \lambda)
\end{align}
It is computationally intractable to directly calculate this posterior
distribution. Therefore, a vibrational distribution $q(\pi, Z, \{\Theta_{k}
\}_{k=1}^{K}, \Theta_{0})$ is used to approximate the posterior
distribution $p$. This distribution $q$ can be factorized such that
\begin{equation}
\label{eq18}
q({\pi }, Z, \{\Theta_{k} \}_{k=1}^{K}, \Theta_{0})=q_{{\pi }}
({\pi)}\prod\limits_{n=1}^N {q_{z_{n} } (z_{n})} \prod\limits_{k=1}^K
{q_{\Theta_{k} } (\Theta_{k})} q_{\Theta_{0} } (\Theta_{0})
\end{equation}
Then estimate each factorization using $W, \varsigma, \eta, \lambda $ and
lastly get the probability of (19).

Lastly, according to the similarity matrix and given hyper-parameters, we
can get the element probability density function of $W_{ij} $, namely, the
value of $p_{0} $ and $p_{2} $.

\section{Complexity analysis}

PCS-SVM method includes three steps: first, developing the first SVM model
through training original data set; then, computing the new $C$ value; last,
using the optimization $C$ value trains the SVM again. The SVM complexity is $O
(dn^{2})$, $d$ is the features dimension and $n$ is the data set size.
Calculation the posterior distribution for the latent variables given the
observed similarity matrix and the hyper-parameters has the complexity of $O
(n)$. Whereas, the complexity of our proposed method is $O (dn^{2})$.

\section{Experiments}

\subsection{Data sets and compared methods}

We tested the proposed algorithm on 16 Keel data sets and our coal and gas
burst monitoring data in first example. We focus on binary-class imbalanced
problem, however, our method can be easily extended to multi-class imbalance
classification. Information about standard benchmark datasets data sets is
summarized in Table~1.

\begin{table}[!t]
\centering
\caption{Data sets used in this article}
\setlength\tabcolsep{6pt}
\begin{tabular}{lccc}
\hline\noalign{\smallskip}
\textbf{Dataset} & Imbalance Ratio& Number of instances& Number of attributes \\
\noalign{\smallskip}
\hline
\noalign{\smallskip}
glass1& 1.82& 214& 9 \\
pima& 1.87& 768& 8 \\
wisconsion& 1.86& 683& 9 \\
haber& 2.78& 306& 3 \\
vehicle0& 3.25& 846& 18 \\
yeast3& 8.1& 1484& 8 \\
ecoli3& 8.6& 336& 7 \\
ecolli$-$2-3-5& 9.17& 244& 7 \\
vowel0& 9.98& 988& 13 \\
ecoli$-$0-1{\_}vs{\_}5& 11& 240& 6 \\
yeast$-$1{\_}vs{\_}7& 14.3& 459& 7 \\
abalone9-18& 16.4& 731& 8 \\
flare-F& 23.79& 1066& 11 \\
yeast4& 28.1& 1484& 8 \\
yeast5& 32.73& 1484& 8 \\
abalone19& 129.44& 4174& 8 \\
\hline
\end{tabular}
\label{tab1}
\end{table}

First, we tested our proposed algorithm in comparison with five other
single classification algorithms based on SVM, including:

\begin{enumerate}[1.]
\item[1.]Standard SVM (SSVM): SVM classifier without any class imbalance
correction technologies.

\item[2.] SVM with random undersampling (SVM-RUS): It firstly adopts RUS to
preprocess the original training set, then trains SVM classifier using the
balanced training data.

\item[3.] SVM with random oversampling (SVM-ROS): It first adopts ROS to preprocess
the original training set, then trains SVM with classifier using the
balanced training data.

\item[4.] SVM with SMOTE (SVM-SMOTE)~\cite{15}: It firstly adopts SMOTE algorithm to
oversampling the minority class instances and to make the data set balance,
then trains SVM classifier using the balanced training data.

\item[5.] Weighted SVM (CS-SVM)~\cite{10} : It assigns different values for the penalty factor
C belong to different classes, to guarantee the fairness of classifier
modeling, we make C$^{+}$/C$^{-}$ equal to imbalance ratio (IR).
\end{enumerate}

In the experiments, for each data set, we performed a fivefold cross
validation. In views of the randomness of classification results, we
repeated the cross-validation process for 20 times and provided the results
in the mean values. All algorithms were implemented in Matlab 2015b running
environment and SVM was realized by libSVM toolbox. Specifically, for SVM,
polynomial kernel and RBF kernel function was adopted, the penalty factor C
and the width parameter $\sigma $ of RBF function were all tuned by using grid
search. We evaluated these six classification algorithms by three mainly
used evaluation criterions in imbalanced classification: F-measure,
G{\_}mean and AUC. For each evaluation criterions and kernel function, we
separate our method into two groups: one group is algorithm-level SVM method, our method PCS-SVM is compared with SVM and CS-SVM; The other group
is sampling-level SVM method, our method combined with SMOTE named as
PCS-SMOTE-SVM is compared with under-sampling(SVM-RUS), oversampling
(SVM-ROS) and SMOTE sampling SVM methods (SVM-SMOTE).

\subsection{Standard benchmark datasets experiment results}

In Table~2 to Table~7, seven SVM methods are evaluated on three measures on
sixteen standard benchmark datasets, and each evaluation measure is
separated into two groups according to two kernel function: Polynomial
kernel (Tables~2, 4, and 6) and RBF kernel (Tables~3, 5, and 7). In the last
row of each table, we list the number of winning method for all sixteen
datasets. In terms of wins number, in general, PCS-SVM and PCS-SMOTE-SVM
method wins more data sets with polynomial kernel than RBF kernel. In
algorithm-level SVM methods, PCS-SVM has the clear superiority on F-measure
and G-means and on the AUS measure. In sampling-level SVM methods,
PCS-SMOTE-SVM outperforms on most of the datasets.  The SVM-SMOTE is close
to PCS-SMOTE-SVM on three measures with RBF kernel.

\begin{table}[h]
\centering
\caption{F-measure values of seven compared single classifiers on 16 Keel
data sets, where bold indicates the best result on each data set.
(\textbf{Polynomial kernel function})}
\setlength\tabcolsep{4.5pt}
\scriptsize\begin{tabular}{lccc|ccccc}
\hline\noalign{\smallskip}
F-measure& SVM& CS-SVM& \textbf{PCS-SVM}& SVM-RUS& SVM-ROS& SVM-SMOTE& \textbf{PCS-SMOTE-SVM} \\
\noalign{\smallskip}
\hline
\noalign{\smallskip}
glass1& 0.5702& 0.5993& \textbf{0.6226} & 0.7597& 0.7283& 0.7649& \textbf{0.7709} \\
pima& 0.3818& \textbf{0.4277}& 0.4233& \textbf{0.5647}& 0.5176& 0.4382& 0.4792 \\
wisconsion& 0.9062& 0.9003& \textbf{0.9212}& 0.9657& 0.9634& 0.9605& \textbf{0.9665} \\
haber& 0.3502& 0.3840& \textbf{0.4437}& 0.5319& 0.5864& 0.6367& \textbf{0.9453} \\
vehicle0& 0.9812& \textbf{0.9847}& 0.9443& 0.9509& 0.9863& 0.9433& \textbf{0.9879} \\
yeast3& 0.7555& 0.7025& \textbf{0.7462}& 0.9077& 0.9395& \textbf{0.9586}& 0.9541 \\
ecoli3& 0.5810& 0.5618& \textbf{0.5937}& 0.7661& 0.9394& 0.9547& \textbf{0.9774} \\
ecolli$-$2-3-5& 0.6294& \textbf{0.6305}& 0.5879& 0.6807& 0.9710& 0.9495& \textbf{0.9836} \\
vowel0& \textbf{0.9848} & 0.9841& 0.9765& 0.9636& 0.9875& \textbf{0.9991}& 0.9988 \\
ecoli$-$0-1{\_}vs{\_}5& 0.7591& 0.7749& \textbf{0.7940}& 0.8581& 0.9882& 0.9905& \textbf{0.9953} \\
yeast$-$1{\_}vs{\_}7& \textbf{0.3742}& 0.2333& 0.2270& 0.5864& 0.8655& \textbf{0.8662}& 0.8511 \\
abalone9-18& 0.4169& 0.4023& \textbf{0.4608}& 0.8255& \textbf{0.8664}& 0.8528& 0.8101 \\
flare-F& 0.1213& \textbf{0.2391}& 0.2390& 0.7199& 0.8180& 0.8954& \textbf{0.9589} \\
yeast4& 0.2422& 0.2957& \textbf{0.3709}& 0.6783& 0.9036& \textbf{0.9272}& 0.9094 \\
yeast5& 0.6009& 0.6363& \textbf{0.6392}& 0.8882& 0.9850& 0.9851& \textbf{0.9864} \\
abalone19& 0.0236& 0.0390& \textbf{0.0540} & 0.7644& 0.7963& \textbf{0.8351}& 0.7430 \\
wins& 2/16& 3/16& \textbf{11/16}& 1/16& 2/16& 5/16& \textbf{9/16} \\
\hline
\end{tabular}
\label{tab2}
\end{table}

\begin{table}[!t]
\centering
\caption{F-measure values of seven compared single classifiers on 16 Keel
data sets, where bold indicates the best result on each data set.
(\textbf{RBF kernel function})}
\centering
\setlength\tabcolsep{4.6pt}
\scriptsize\begin{tabular}{lccc|ccccc}
\hline\noalign{\smallskip}
\textbf{F-measure} & SVM& CS-SVM& \textbf{PCS-SVM}& SVM-RUS& SVM-ROS& SVM-SMOTE& \textbf{PCS-SMOTE-SVM} \\
\noalign{\smallskip}
\hline
\noalign{\smallskip}
glass1& 0.4678& 0.5662& \textbf{0.6721}& 0.8351& \textbf{0.8571}& 0.8463& 0.8416 \\
pima& 0.5032& 0.4897& \textbf{0.6274}& 0.7568& 0.7555& 0.7717& \textbf{0.7570} \\
wisconsion& 0.9495& \textbf{0.9501}& 0.9484& \textbf{0.9762}& 0.9686& 0.9682& 0.9684 \\
haber& 0.1458& 0.3953& \textbf{0.4187}& 0.5860& 0.7206& 0.7092& \textbf{0.9195} \\
vehicle0& \textbf{0.9386}& 0.9250& 0.9253& 0.7261& \textbf{0.9395}& 0.9226& 0.9272 \\
yeast3& \textbf{0.7698}& 0.7000& 0.7606& 0.9124& 0.9435& \textbf{0.9620}& 0.9580 \\
ecoli3& 0.6337& 0.6225& \textbf{0.6658}& 0.8404& 0.8344& \textbf{0.9551}& 0.9512 \\
ecolli$-$2-3-5& \textbf{0.7452} & 0.7562& 0.6000& 0.7544& \textbf{0.9735}& 0.7792& 0.9412 \\
vowel0& 0.9772& 0.9853& \textbf{0.9939}& 0.9691& 0.8794& 0.9882& \textbf{0.9994} \\
ecoli$-$0-1{\_}vs{\_}5& 0.7383& 0.7589& \textbf{0.7792}& 0.7581& \textbf{0.9924}& 0.7453& 0.6749 \\
yeast$-$1{\_}vs{\_}7& \textbf{0.4712}& 0.2632& 0.2596& 0.6518& 0.8653& 0.8606& \textbf{0.8670} \\
abalone9-18& 0.2487& 0.4802& \textbf{0.5125}& 0.6699& \textbf{0.9529}& 0.9092& 0.8864 \\
flare-F& 0.2134& 0.2726& \textbf{0.2895}& 0.7822& 0.8819& 0.9291& \textbf{0.9340} \\
yeast4& 0.3021& 0.3323& \textbf{0.4201}& 0.7886& 0.8043& \textbf{0.9373}& 0.9138 \\
yeast5& 0.6821& 0.6855& \textbf{0.7176}& 0.8580& 0.9900& \textbf{0.9904}& 0.9892 \\
abalone19& 0.0220& \textbf{0.0223}& 0.1027& 0.4563& 0.7404& 0.0229& \textbf{0.7405} \\
wins& 4/16& 2/16& \textbf{10/16}& 1/16& 5/16& 4/16& \textbf{6/16} \\
\hline
\end{tabular}
\label{tab3}


\centering
\caption{G-mean values of seven compared single classifiers on 16 Keel data
sets, where bold indicates the best result on each data set.
(\textbf{Polynomial kernel function})}
\centering
\setlength\tabcolsep{4.6pt}
\scriptsize\begin{tabular}{lccc|ccccc}
\hline\noalign{\smallskip}
G-mean & SVM& CS-SVM& \textbf{PCS-SVM}& SVM-RUS& SVM-ROS& SVM-SMOTE& \textbf{PCS-SMOTE-SVM} \\
\noalign{\smallskip}
\hline
\noalign{\smallskip}
glass1& 0.6597& 0.6592& \textbf{0.6660}& \textbf{0.7711}& 0.7157& 0.7509& 0.7626 \\
pima& 0.4213& 0.4905& \textbf{0.4931}& 0.4113& 0.5067& 0.5105& \textbf{0.5241} \\
wisconsion& 0.9230& 0.9150& \textbf{0.9374}& \textbf{0.9693}& 0.9623& 0.9604& 0.9629 \\
haber& 0.5072& 0.5209& \textbf{0.6859}& 0.5328& 0.5721& 0.6293& \textbf{0.9669} \\
vehicle0& \textbf{0.9682}& 0.9616& 0.9633& 0.9500& 0.9860& 0.9644& \textbf{0.9864} \\
yeast3& 0.8495& \textbf{0.9045}& 0.8886& 0.9048& 0.9378& \textbf{0.9578}& 0.9515 \\
ecoli3& 0.7667& 0.7842& \textbf{0.7999}& 0.7800& 0.9357& 0.9524& \textbf{0.9552} \\
ecolli$-$2-3-5& 0.8173& \textbf{0.8542}& 0.8468& 0.7056& 0.9668& 0.9465& \textbf{0.9958} \\
vowel0& \textbf{0.9932} & 0.9813& 0.9899& 0.9619& 0.9723& 0.9881& \textbf{0.9988} \\
ecoli$-$0-1{\_}vs{\_}5& 0.8755& 0.8534& \textbf{0.8831}& 0.8794& 0.9878& 0.9905& \textbf{0.9954} \\
yeast$-$1{\_}vs{\_}7& 0.5334& 0.6136& \textbf{0.7221}& 0.5878& \textbf{0.8512}& 0.8511& 0.7989 \\
abalone9-18& 0.6681& 0.8223& \textbf{0.8292}& 0.8214& \textbf{0.8647}& 0.8561& 0.8480 \\
flare-F& 0.4347& \textbf{0.7108}& 0.6976& 0.7310& 0.8166& 0.9005& \textbf{0.9573} \\
yeast4& 0.3987& 0.7474& \textbf{0.7498}& 0.6812& 0.9188& \textbf{0.9257}& 0.8981 \\
yeast5& 0.8042& 0.8415& \textbf{0.8517}& 0.8883& 0.9807& 0.9813& \textbf{0.9865} \\
abalone19& 0& 0.7190& \textbf{0.7808} & 0.7853& 0.7820& \textbf{0.8309}& 0.7962 \\
wins& 2/16& 3/16& \textbf{11/16}& 2/16& 2/16& 3/16& \textbf{9/16} \\
\hline
\end{tabular}
\label{tab4}
\end{table}

\begin{table}[!t]
\centering
\caption{G-mean values of seven compared single classifiers on 16 Keel data
sets, where bold indicates the best result on each data set. (\textbf{RBF
kernel function})}
\centering
\setlength\tabcolsep{4.6pt}
\scriptsize\begin{tabular}{lccc|ccccc}
\hline\noalign{\smallskip}
\textbf{G-mean} & SVM& CS-SVM& \textbf{PCS-SVM}& SVM-RUS& SVM-ROS& SVM-SMOTE& \textbf{PCS-SMOTE-SVM} \\
\noalign{\smallskip}
\hline
\noalign{\smallskip}
glass1& 0& 0.6610& \textbf{0.6721}& 0.8407& \textbf{0.8569}& 0.8459& 0.8161 \\
pima& 0& 0.1965& \textbf{0.6995}& 0.6183& 0.7805& \textbf{0.7928}& 0.7806 \\
wisconsion& 0.9679& \textbf{0.9696}& 0.9683& \textbf{0.9749}& 0.9676& 0.9673& 0.9667 \\
haber& 0.2809& 0.5614& \textbf{0.5653}& 0.6169& 0.7159& 0.7082& \textbf{0.9535} \\
vehicle0& 0.9574& 0.9579& \textbf{0.9641}& 0.4984& 0.9213& 0.9265& \textbf{0.9298} \\
yeast3& 0.8507& \textbf{0.8921}& 0.8861& 0.9107& 0.9420& \textbf{0.9612}& 0.9555 \\
ecoli3& 0.7714& 0.8702& \textbf{0.8797}& 0.8483& 0.9304& 0.9537& \textbf{0.9698} \\
ecolli$-$2-3-5& \textbf{0.8454} & 0& 0.8308& 0.7800& 0.9743& 0.8003& \textbf{0.9844} \\
vowel0& 0.9800& 0.9950& \textbf{0.9988}& 0.9678& 0.9788& 0.9801& \textbf{0.9995} \\
ecoli$-$0-1{\_}vs{\_}5& 0.8676& 0& \textbf{0.8739}& 0.8496& \textbf{0.8925}& 0.7711& 0.7696 \\
yeast$-$1{\_}vs{\_}7& 0.5698& 0.6452& \textbf{0.6560}& 0.6533& 0.8565& \textbf{0.8654}& 0.8248 \\
abalone9-18& 0.3756& 0.8301& \textbf{0.8330}& 0.6979& \textbf{0.9105}& 0.9089& 0.8683 \\
flare-F& 0& \textbf{0.8295}& 0.8068& 0.7798& 0.8699& \textbf{0.9346}& 0.9283 \\
yeast4& 0.2019& 0.7821& \textbf{0.7861}& 0.7834& 0.9094& 0.9058& \textbf{0.9132} \\
yeast5& 0.8198& 0.8463& \textbf{0.8720}& 0.8584& 0.9728& \textbf{0.9905}& 0.9893 \\
abalone19& 0& \textbf{0.6000}& 0.3720& 0.2169& 0.6275& \textbf{0.6477}& 0.5685 \\
wins& 1/16& 4/16& \textbf{11/16}& 1/16& 3/16& \textbf{6/16}& \textbf{6/16} \\
\hline
\end{tabular}
\label{tab5}


\caption{AUS values of seven compared single classifiers on 16 Keel data
sets, where bold indicates the best result on each data set.
(\textbf{Polynomial kernel function})}
\centering
\setlength\tabcolsep{4.6pt}
\scriptsize\begin{tabular}{lccc|ccccc}
\hline\noalign{\smallskip}
\textbf{AUS} & SVM& CS-SVM& \textbf{PCS-SVM}& SVM-RUS& SVM-ROS& SVM-SMOTE& \textbf{PCS-SMOTE-SVM} \\
\noalign{\smallskip}
\hline
\noalign{\smallskip}
glass1& 0.2499& 0.1956& \textbf{0.6721}& 0.1584& 0.8089& \textbf{0.8334}& 0.8103 \\
pima& 0.5678& 0.6019& \textbf{0.6375}& 0.4779& 0.5751& 0.6347& \textbf{0.6600} \\
wisconsion& 0.5610& \textbf{0.5707}& 0.5465& 0.5104& 0.9754& 0.9756& \textbf{0.9767} \\
haber& 0.5031& \textbf{0.5354}& 0.5674& 0.5206& 0.6186& 0.6776& \textbf{0.9918} \\
vehicle0& \textbf{0.8257}& 0.6626& 0.6629& 0.0182& 0.9955& 0.9901& \textbf{0.9959} \\
yeast3& 0.0354& 0.0384& \textbf{0.0430}& 0.0451& 0.9789& \textbf{0.9877}& 0.9714 \\
ecoli3& 0.5061& \textbf{0.5812}& 0.4998& 0.1505& 0.9577& 0.9740& \textbf{1} \\
ecolli$-$2-3-5& 0.5101& 0.5597& \textbf{0.5995}& 0.6344& 0.9782& 0.9758& \textbf{1} \\
vowel0& \textbf{0.6667} & 0.6664& 0.6665& 0.7282& 0.9555& 0.9657& \textbf{0.9997} \\
ecoli$-$0-1{\_}vs{\_}5& \textbf{0.5672} & 0.5663& 0.5533& 0.5644& 0.9966& \textbf{0.9972}& 0.8977 \\
yeast$-$1{\_}vs{\_}7& 0.2433& \textbf{0.2810}& 0.2761& 0.3464& 0.8999& \textbf{0.9183}& 0.8716 \\
abalone9-18& 0.4795& 0.5902& \textbf{0.5967}& 0.5913& 0.9480& 0.9300& \textbf{0.9679} \\
flare-F& 0.5457& \textbf{0.5658}& 0.5416& 0.2308& 0.8907& 0.9632& \textbf{0.9688} \\
yeast4& 0.1167& 0.1253& \textbf{0.1588}& 0.2396& 0.9672& 0.9703& \textbf{0.9745} \\
yeast5& 0.5376& 0.6399& \textbf{0.6415}& 0.6558& 0.9903& 0.9914& \textbf{0.9934} \\
abalone19& \textbf{0.4885}& 0.4840& 0.4785& 0.4926& 0.8479& \textbf{0.8904}& 0.8691 \\
wins& 4/16& 5/16& \textbf{7/16}& 0/16& 0/16& 5/16& \textbf{11/16} \\
\hline
\end{tabular}
\label{tab6}
\end{table}

\begin{table}[!t]
\caption{AUS values of seven compared single classifiers on 16 Keel data
sets, where bold indicates the best result on each data set. (\textbf{RBF
kernel function})}
\centering
\setlength\tabcolsep{4.6pt}
\scriptsize\begin{tabular}{lccc|cccccc}
\hline\noalign{\smallskip}
\textbf{AUS} & SVM& CS-SVM& \textbf{PCS-SVM}& SVM-RUS& SVM-ROS& SVM-SMOTE& \textbf{PCS-SMOTE-SVM} \\
\noalign{\smallskip}
\hline
\noalign{\smallskip}
glass1& 0& 0.4576& \textbf{0.6721}& 0.0941& \textbf{0.9133}& 0.9104& 0.9111 \\
pima& 0& 0.6113& \textbf{0.6132}& 0.2358& 0.8592& \textbf{0.9289}& 0.9048 \\
wisconsion& 0.0097& 0.0108& \textbf{0.0117}& 0.0037& 0.9907& 0.9901& \textbf{0.9909} \\
haber& 0.4374& \textbf{0.5506}& 0.5337& 0.5349& 0.8092& 0.7830& \textbf{0.9916} \\
vehicle0& 0.6656& 0.7737& \textbf{0.7856}& 0.0555& 0.9955& 0.9927& \textbf{0.9966} \\
yeast3& 0.5253& \textbf{0.5338}& 0.5325& 0.0380& 0.9809& 0.9883& \textbf{0.9889} \\
ecoli3& \textbf{0.5885} & 0.4481& 0.5497& 0.5812& 0.9680& \textbf{0.9791}& 0.9786 \\
ecolli$-$2-3-5& 0.5599& \textbf{0.6052}& 0.5880& 0.6393& 0.7765& 0.7989& \textbf{1} \\
vowel0& 0.6667& 0.6667& \textbf{0.6680}& 0.4121& 0.9776& 0.9881& \textbf{1} \\
ecoli$-$0-1{\_}vs{\_}5& 0.5698& 0.5589& \textbf{0.5742}& 0.6764& 0.9989& 0.7814& \textbf{1} \\
yeast$-$1{\_}vs{\_}7& \textbf{0.2549}& 0.2448& 0.2512& 0.2593& 0.9326& \textbf{0.9424}& 0.9201 \\
abalone9-18& \textbf{0.5711}& 0.4877& 0.4891& 0.6485& \textbf{0.9838}& 0.9680& 0.9618 \\
flare-F& \textbf{0.6432}& 0.5243& 0.5127& 0.7026& 0.9494& 0.9543& \textbf{0.9863} \\
yeast4& 0.1668& 0.1675& \textbf{0.1775}& 0.1358& 0.9762& \textbf{0.9768}& 0.9745 \\
yeast5& 0.4325& 0.4372& \textbf{0.4392}& 0.5554& \textbf{0.9975}& 0.9969& 0.9967 \\
abalone19& \textbf{0.4527}& 0.2917& 0.3219& 0.3078& 0.7807& \textbf{0.8967}& 0.8609 \\
wins& 5/16& 3/16& \textbf{8/16}& 0/16& 3/16& 5/16& \textbf{8/16} \\
\hline
\end{tabular}
\label{tab7}
\end{table}

\begin{figure}[ht!]
\centering
\subfigure[]{\includegraphics[width=.486\textwidth]{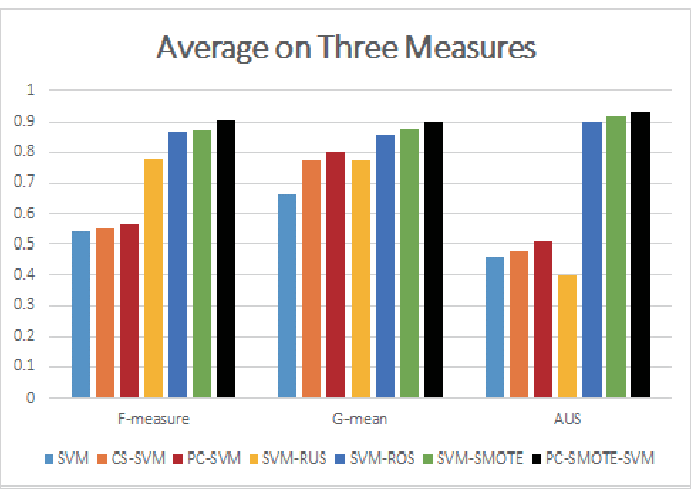}}\hfill
\subfigure[]{\includegraphics[width=.486\textwidth]{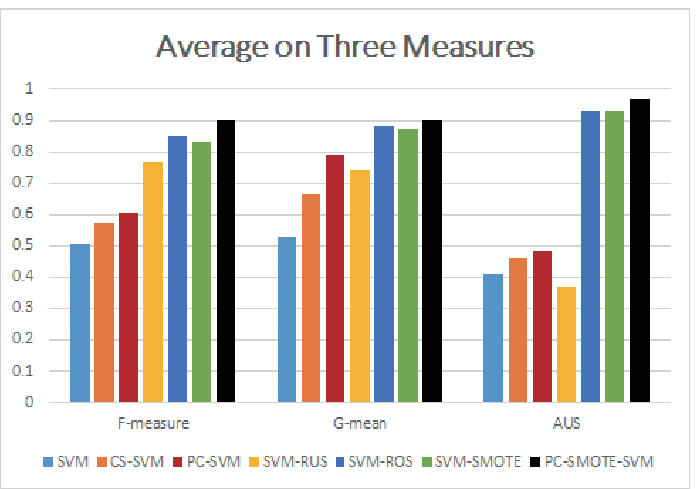}}
\caption{(a) Average on Three Measures \textbf{(Polynomial kernel function)}; (b) Average on Three Measures \textbf{(RBF kernel function)}}
\label{fig3}
\end{figure}

We compute the average evaluation measure for each SVM method on all
datasets, and the results is shown in Fig.~3. The red and black bar
represent our proposed method PCS-SVM and PCS-SMOTE-SVM respectively. We can
conclude that two kernel function has no obvious influence on average on
three measures for all methods. PCS-SVM is the best algorithm in
algorithm-level SVM group and for the G-means measure, PCS-SVM even better
than SVM-RUS. Four sampling-level algorithms all have the excellent
performance in AUS measure and F-measure. PCS-SMOTE-SVM has the best
performance in all sixteen date sets and all the average value on three
measures are above or very close to 0.9.

\subsection{Coal and gas burst monitoring data set experiment results}

In this section, we evaluate PCS-SVM and PCS-SMOTE-SVM on our real-world data
set: coal and gas burst electromagnetic monitoring data, as shown in Fig.~1. Coal and gas burst is a kind of damaging accident in coal mining.
Electromagnetic data can effective reflect the premonition of coal and gas
burst accident. So if we can find the premonition of burst accident from the
electromagnetic data accurately and timely, we can pre-alarming the burst
accident. In our example, the data set is one month electromagnetic
monitoring data with the number of 7989, and there are only 210 premonition
data, with the imbalance ratio of 37.04. The data set has two features:
electromagnetic intensity and electromagnetic pulse. We set the premonition
data with positive class and the other normal data is negative class. We
expect the premonition data can be classified from the normal data and
higher accuracy of these premonition data than normal data. Because negative
class is so huge, we random undersampling negative data to 2000 and SMOTE
positive data to 500. As is concluded in Section~4.2, SMOTE has the better
performance than ROS and RUS method, and PCS-SVM and PCS-SMOTE-SVM method wins
more data sets with polynomial kernel than RBF kernel, we only use SMOTE to
oversampling data and adopt polynomial kernel. In this experiment, besides
three evaluations we have used, the sensitivity and specificity also be
evaluated in order to describe the accuracy rate of positive class. The results are shown in Table~8.

\begin{table}
\caption{The evaluate results on electromagnetic monitoring data}
\centering
\setlength\tabcolsep{6pt}
\begin{tabular}{llllll}
\hline\noalign{\smallskip}
& Sensitivity& Specificity& F-measure& G-mean& AUC \\
\noalign{\smallskip}
\hline
\noalign{\smallskip}
SVM& 0.0211& 0.9722& 0.0379& 0.1432& 0.4094 \\
CS-SVM& 0.6599& 0.3509& 0.2670& 0.2146& 0.4923 \\
SMOTE-SVM& 0.5445& 0.4575& 0.2820& 0.4136& 0.4816 \\
\textbf{PCS-SVM}& 0.6055& 0.3910& 0.3104& 0.2244& 0.4998 \\
\textbf{PCS-SMOTE-SVM}& 0.9935& 0.1411& 0.3713& 0.4036& 0.7770 \\
\hline
\end{tabular}
\label{tab8}
\end{table}

From the Table~8, we can easily conclude that both PCS-SVM and PCS-SMOTE-SVM
outperform compared method. Especially, sensitivity is higher than
specificity which attain our expectation. PCS-SMOTE-SVM has the best
performance on all the evaluation measures and the AUC values reaches
0.7770, compared to SMOTE-SVM, improve 29.54{\%}.

\section{Conclusion and future work}

This paper presents  a novel C value optimization method on the basis of
cluster probability density function, which is estimated only using
similarity matrix and some predefined hyper-parameters. Experimental results
on various standard benchmark datasets and real-world data with different
ratios of imbalance show that the proposed method is effective in comparison
with commonly used cost-sensitive techniques.

For future work, we plan to leverage multi-view feature representations ~\cite{16,17,18,19} to improve the performance of SVMs for imbalanced data set.

\end{document}